\documentclass[letterpaper]{article}

\usepackage{natbib,alifeconf}  %% The order is important
\usepackage{url}
\usepackage[hidelinks]{hyperref}%,cleveref}
\usepackage{booktabs}

\usepackage{verbatim}
\usepackage{todonotes}
\usepackage[shortlabels]{enumitem}
\usepackage{etoolbox}
\usepackage{subfig}
\usepackage{float}
\usepackage{amsmath}
\usepackage{cleveref}

\newtoggle{ShowNames}
\toggletrue{ShowNames}
\setlist[itemize]{noitemsep}

\iftoggle{ShowNames}{
  \author{Matias Barandiaran \and James Stovold \\
  \mbox{} \\
  Lancaster University Leipzig,
  Nikolaistra\ss{}e 10,
  04109 Leipzig, Germany \\
  Email: \url{j.stovold@lancaster.ac.uk}}%
}{ %
  \author{A. N. Author$^1$ \\
  \mbox{} \\
  $^1$Address Line 1 \\
  Email: \url{a.n.author@institution.com} }
}

\title{Growing Reservoirs with Developmental Graph Cellular Automata}

\begin{document}

\maketitle

\begin{abstract}

    Developmental Graph Cellular Automata (DGCA) are a novel model for morphogenesis, capable of growing directed graphs from single-node seeds. In this paper, we show that DGCAs can be trained to grow reservoirs. Reservoirs are grown with two types of targets: task-driven (using the NARMA family of tasks) and task-independent (using reservoir metrics).

    Results show that DGCAs are able to grow into a variety of specialized, life-like structures capable of effectively solving benchmark tasks, statistically outperforming `typical' reservoirs on the same task. Overall, these lay the foundation for the development of DGCA systems that produce plastic reservoirs and for modeling functional, adaptive morphogenesis.

\end{abstract}

% Choose one of: Full Paper, Summaries, or Late Breaking Abstracts 
% \noindent Submission type: \textbf{Full Paper}\\
% If sharing code / data, anonymize your repository and paste the link here.
% Example of anonymizing sevice for github: https://anonymous.4open.science/
% delete this line if not needed    
\noindent Code available at: \iftoggle{ShowNames}{\url{https://github.com/mbaranr/alife2025}}{\url{https://anonymous.4open.science/r/alife2025}}

\begin{comment}

Story: Focus on growth
 - Growth is a core mechanism in life/nature
 - Morphogenesis is the growth of a particular shape
 - NCAs give us a shape that we have chosen as a target, grown from a seed cell
 - Nature, however, grows because it needs something or to produce something (i.e. there is a purpose to it growing something)
 - DGCAs allows us to grow graphs based on a fitness function (i.e. we get growth with a purpose)

- the particular purpose we have chosen is computing
- typical approach to RC is basically brute-force
- while some have tried using optimization algorithms, most of these are just optimising the brute-force approach
- when it comes to PRC, brute-force is often the only choice (find a blob, test the blob, etc.)
- why don't we use nature's approach and grow a reservoir instead? 

\end{comment}

\section{Introduction}

Morphogenesis orchestrates the development of complex life. Through morphogenesis, populations of cells collaborate to form intricate structures such as eyes, limbs, and hearts. \citet{mordvintsev2020growing} introduced Neural Cellular Automata (NCA) as a model of morphogenesis. By leveraging neural networks as the transition rule of a cellular automata, NCAs can learn to grow predefined two-dimensional shapes from a single-cell seed. This approach, however, lacks an important biological nuance: in nature, structures do not emerge solely for their shape but to serve a purpose. In this study, that purpose is computation.

Developmental Graph Cellular Automata (DGCA) \citep{waldegrave2023developmental} extend NCAs by enabling the growth of directed graphs based on fitness functions. Directed graphs are particularly well-suited for representing recurrent neural networks (RNNs), which excel at modeling sequential data and have become powerful tools in tasks such as time series prediction and natural language processing \citep{mienye2024recurrent}. Despite their utility, training RNNs is computationally intensive due to their reliance on backpropagation through time. Reservoir Computing (RC) was developed as a response to these challenges \citep{lukovsevivcius2009reservoir}. In the RC framework, a large, fixed, randomly connected RNN---known as the reservoir---provides rich nonlinear dynamics while only the readout layer is trained.

A key appeal of RC is its ability to leverage physical reservoirs. This scheme's flexibility allows for a wide range of dynamical systems exhibiting the echo state property to serve as successful reservoirs \citep{nakajima2020physical}; including cultured biological neurons \citep{sumi_biological_2023}, nanomaterial probes \citep{qi_physical_2023}, or even an octopus arm replica \citep{nakajima_soft_2013}. Attention toward Physical Reservoir Computing (PRC) over conventional von Neumann architectures reflects a shift to exploiting the intrinsic dynamics of physical systems. PRC enables low-power computation, continuous signal processing, and robust fault tolerance in ways that traditional architectures struggle to achieve \citep{yan2024emerging}.

Conventional RC relies on brute-force methods, i.e.\ randomly initializing multiple reservoirs in turn until the appropriate dynamics are found. The search is often done randomly or by novelty-based methods. While some work has applied optimization algorithms \citep{soltani2023echo}, they often serve only to refine this brute-force approach. In PRC, the challenge is even greater. Physical substrates are often selected based on general computational potential \citep{dale_substrate-independent_2019}, with little opportunity to actively design or adapt the system.
    
This raises a fundamental question: why not use DGCAs to grow our own reservoirs? Previous applications of DGCAs have focused on growing graphs with specific structural features, such as motif distributions \citep{waldegrave_creating_2024}. In this work, we grow directed graphs based on task performance as well as task-independent metrics. This lays the foundation to develop DGCA systems capable of producing \textit{plastic} reservoirs. Plastic reservoirs would be able to adapt to changes in the environment or task, and recover functionality after sustaining damage. We also find that specialization emerges naturally through task-driven growth. Rather than relying on a ``one-size-fits-all’’ reservoir, we can grow reservoirs that prioritize certain RC metrics without explicitly optimizing for them. This represents an important step toward modeling a more functional, adaptive morphogenesis.

\section{Developmental Graph Cellular Automata}

Conventional Cellular Automata (CA) consist of a grid of cells, each of which can be dead or alive. Neural Cellular Automata (NCA) \citep{mordvintsev2020growing} extend the CA framework by replacing the rule-based update mechanism with a neural network trained to evolve the cell states so that a desired overall pattern emerges from an initial single-cell seed.

Biology, however, does not work on a grid. To better capture the complexity of biological environments, Graph Cellular Automata (GCA) generalize the traditional CA framework by organizing cells in a graph structure, where nodes represent cells and edges represent connections between them. Despite providing a more flexible environment, efforts put into learning GCAs \citep{grattarola2021learning} have focused on self-organization rather than growth (i.e.\ geometric point cloud reconstruction with a fixed number of nodes). 

\cite{waldegrave2023developmental} introduced Developmental Graph Cellular Automata (DGCA) as an extension to NCAs, GCAs, and L-Systems. DGCAs allow for a graph to start from a single node and expand over time. Using information about their own and neighboring states, nodes can decide when to divide, controlling the growth process in a distributed way. The work in this paper builds upon a more recent version of the model, DGCA-M \citep{waldegrave_creating_2024}, which separates the update step into two distinct stages: \textit{action} and \textit{state}. Each requires its own Single-Layer Perceptron (SLP) to perform the transition rule. A similar approach is taken by Neural Developmental Programs \citep{najarro2023towards}, which have been used to grow artificial neural networks from seed graphs.

For a particular graph, the DGCA follows the following steps (visualised in Figure \ref{fig:dgca_pipe}) at each timestep to determine how the graph should change. 

\begin{figure}
  \centering
  \subfloat[From existing]{
    \includegraphics[height=0.2\textwidth]{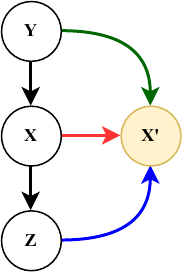}
  }
    \subfloat[To existing]{
    \includegraphics[height=0.2\textwidth]{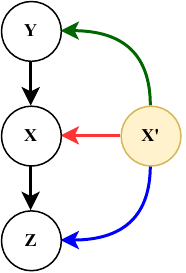}
  }
    \subfloat[To new]{
    \includegraphics[height=0.2\textwidth]{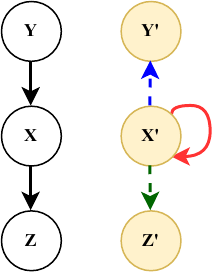}
  }
  \caption{Possible edge choices for a newly created node \(\mathbf{X}'\) resulting from division of \(\mathbf{X}\). Adapted from Figure~6 of \cite{waldegrave_creating_2024}.}
  \label{fig:edges}
\end{figure}

\begin{enumerate}
    \item A neighborhood aggregation function is used to capture both the number and states of neighboring nodes. 
    \item The resulting neighborhood matrix is  passed to the action SLP, which determines each node’s behavior: removal, division, or stasis. All decisions are made at the level of individual nodes, based exclusively on local information.
    \item The graph is restructured according to the valid edge choices for newly created nodes shown in Figure~\ref{fig:edges}. These are grouped into three categories: \textit{From Existing}, \textit{To Existing} and \textit{To New}. In order to prevent the graph from becoming overly dense, nodes are only allowed to select one set of edges from within each category. 
    \item Once the graph topology is updated, the neighborhood matrix is recomputed. This updated matrix is then passed through the state SLP, which determines the new internal states of the nodes for the next timestep.
\end{enumerate}

\section{Methods}

\begin{figure*}
    \centering
    \includegraphics[width=\textwidth]{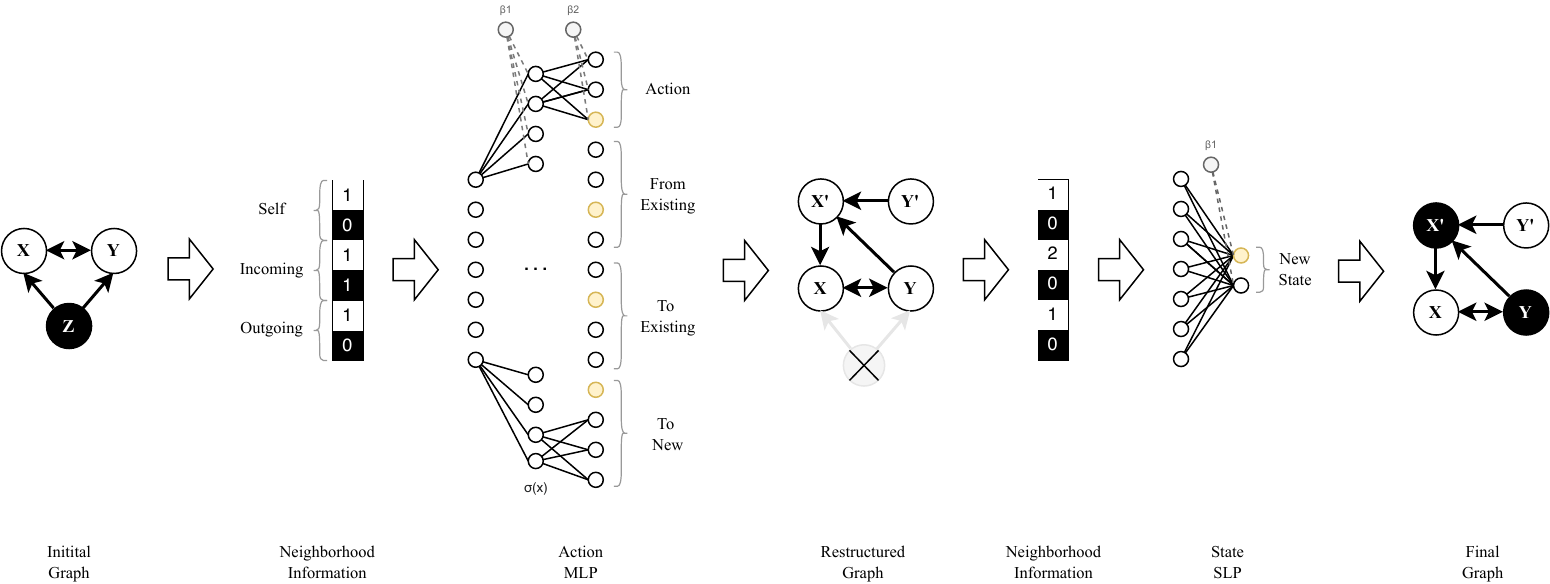}
    \caption{DGCA pipeline of a two-state (black and white) system. The neighborhood information vector \(\mathbf{G}\) is shown for node \(\mathbf{X}\). The action MLP determines that \(\mathbf{X}\) and \(\mathbf{Y}\) divide, while \(\mathbf{Z}\) is removed. \(\mathbf{G}\) is gathered a second time, now with zero incoming black node connections but an additional white one, before being passed to the state SLP. Adapted from Figure~4 of \cite{waldegrave_creating_2024}.}
    \label{fig:dgca_pipe}
\end{figure*}

As many of the most interesting rules in conventional Cellular Automata (CA) are not linearly separable, we extend the DGCA-M model by replacing the `action' Single-Layer Perceptron (SLP) with a 2-layer Multi-Layer Perceptron (MLP) with 64 nodes in the hidden layer (see Figure~\ref{fig:dgca_pipe}). Additionally, nodes that emerge with a degree of zero (isolated) are dropped.

\begin{figure}
    \centering
    \includegraphics[width=\linewidth]{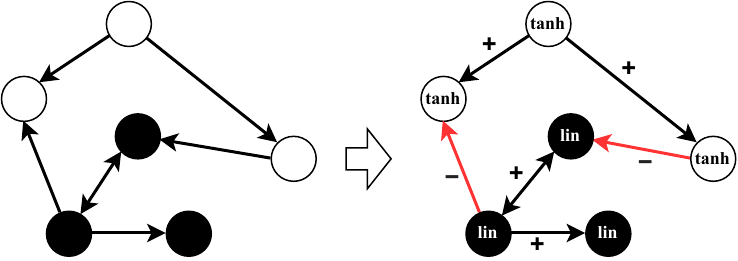}
    \caption{Bipolarization step of a two-state system. Weights between black and white states are negative (\(-1\)). Weights between equal states are positive (\(+1\)). White state nodes represent hyperbolic tangent neurons, while black state nodes represent linear neurons.}
    \label{fig:polar}
\end{figure}

Graphs are bipolarized to enable more complex dynamics, with edge weights assigned values of either $+1$ or $-1$ (see Figure~\ref{fig:polar}). Each weight is determined by the direction of the edge and the states of the connected nodes. This extends the original DGCA framework, which produces graphs with binary weights. Node states are also mapped to activation functions (e.g.\ tanh, sigmoid, linear), allowing for multiple types of neuron behavior. This study uses three-state systems with linear and hyperbolic tangent neurons.

\subsection{Search}

Since the error function is non-differentiable, algorithms like backpropagation are not viable. Instead, we use a variant of the classic Microbial Genetic Algorithm (MGA) \citep{harvey_microbial_2011} to search for the weights. 

Each individual comprises two chromosomes: MLP and SLP weights. To preserve internal dependencies and minimize fragmentation risk, the MLP is encoded as a single chromosome rather than split by layer. In each trial, the algorithm randomly selects two competing individuals. The loser receives 50\% of the winner's genetic material and undergoes a 2\% mutation.

We initialize a population of 10 individuals per chromosome type, resulting in 100 ``genetically complete'' individuals. Due to stochastic selection, not all individuals are guaranteed to be evaluated. For further details on the MGA, the reader is referred to \citet{harvey_microbial_2011}.

\subsection{Experimental Design}

Each experiment consists of 150 runs, where each run involves 1000 iterations of the MGA. Within each iteration (trial) two reservoirs compete against each other. This results in a total of at least $150 \times 1000 \times 2 = 300{,}000$ reservoir instances (not necessarily unique) generated per experiment. To grow each reservoir, the DGCA update pipeline is executed 100 times, starting from a single node seed. 

\begin{figure*}
  \centering
  \subfloat[Linear]{
    \begin{minipage}{0.3\textwidth}
      \centering
      \includegraphics[width=0.45\textwidth]{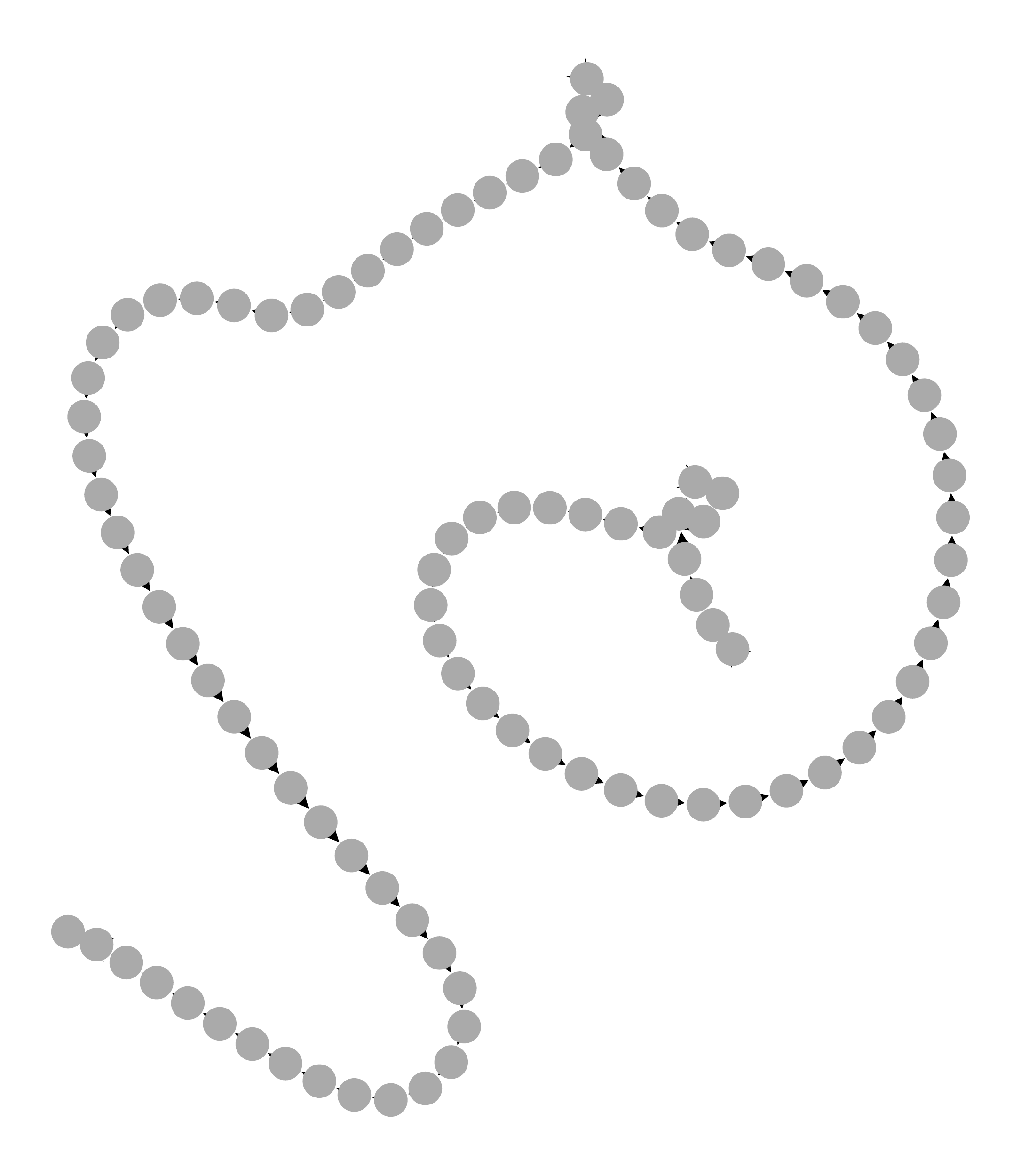}
      \includegraphics[width=0.45\textwidth]{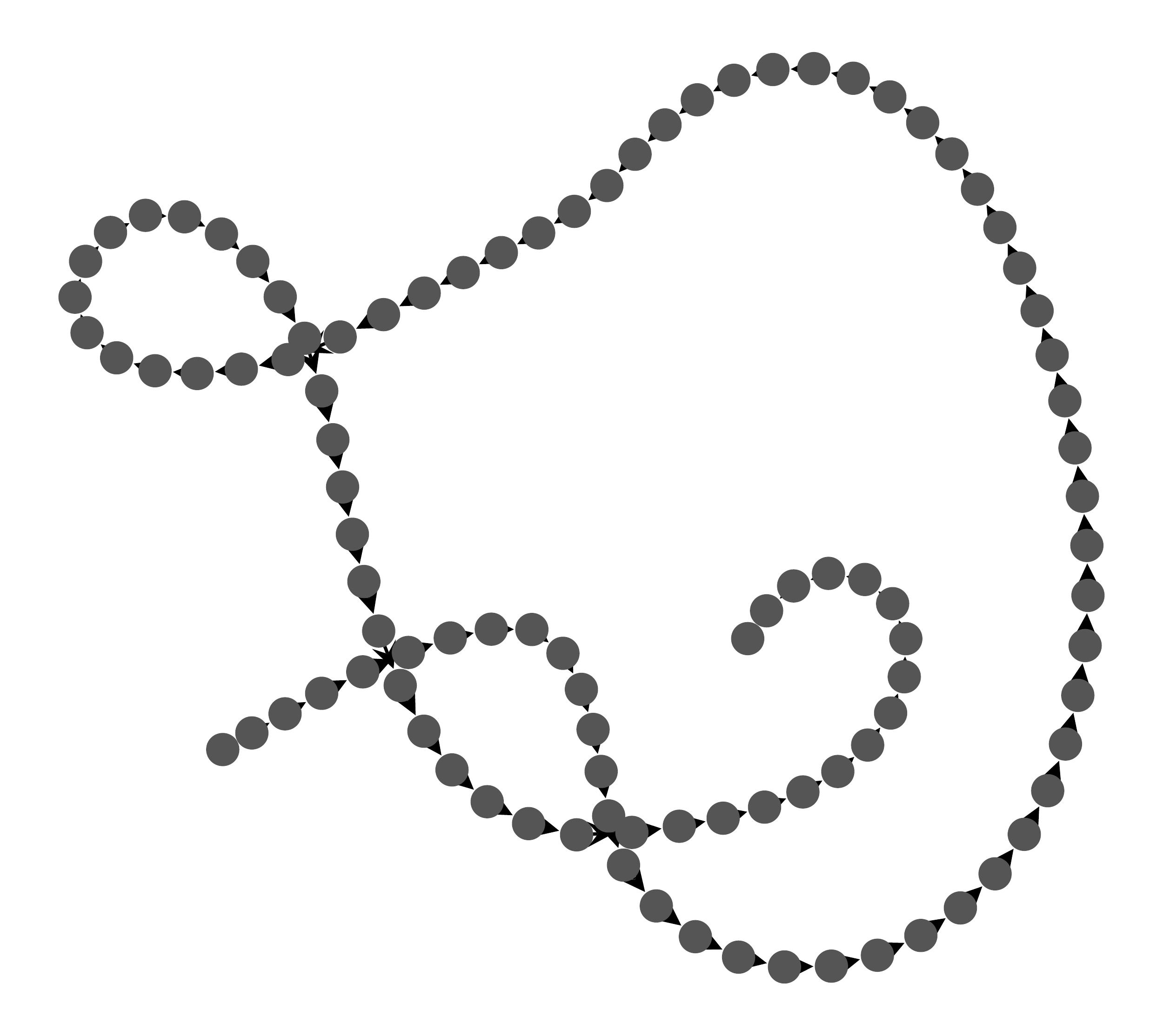}\\
      \includegraphics[height=0.45\textwidth]{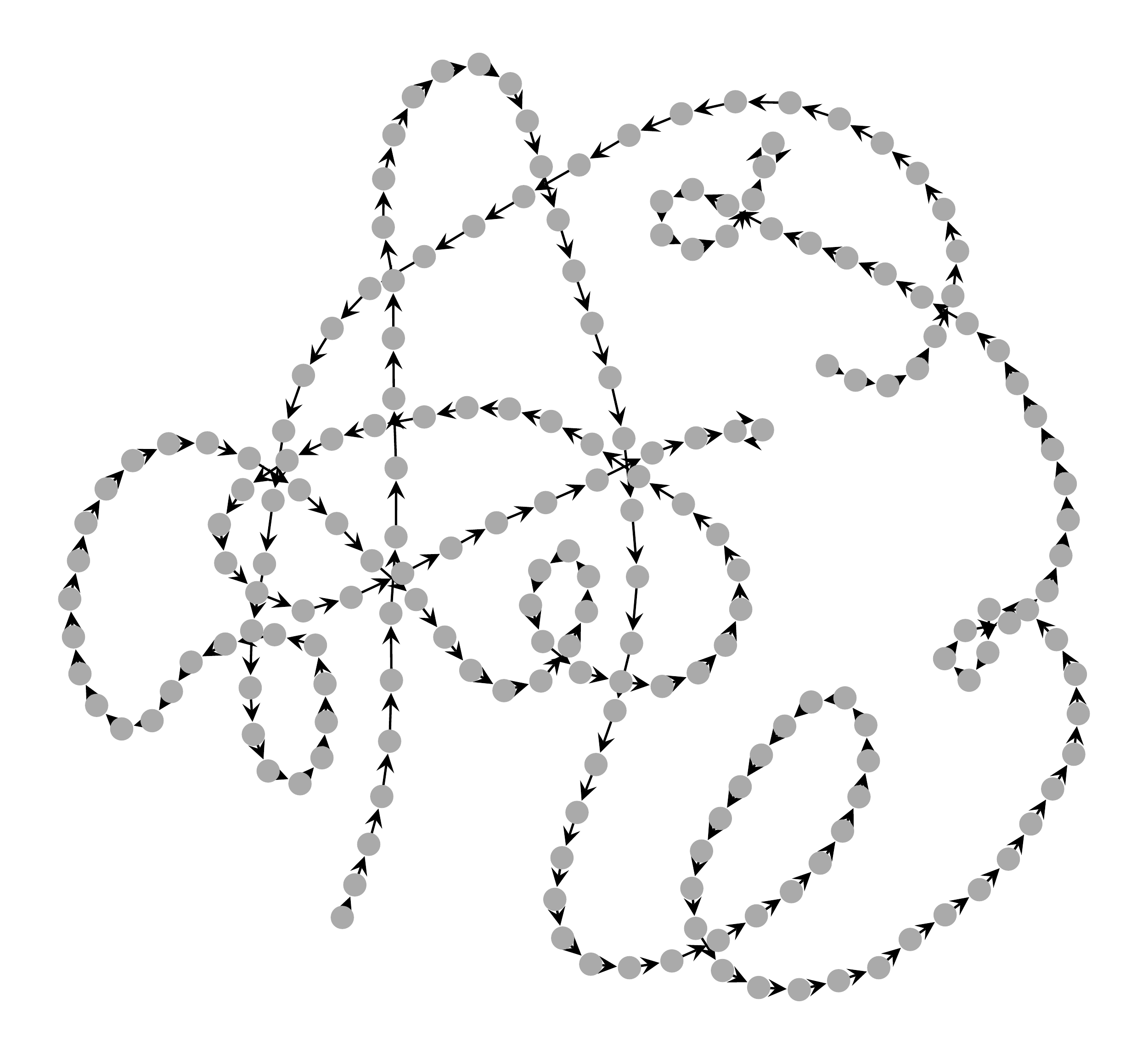}
    \end{minipage}
  }
  \hspace{0.4cm}
  \subfloat[Loosely Stranded]{
    \begin{minipage}{0.3\textwidth}
      \centering
      \includegraphics[width=0.45\textwidth]{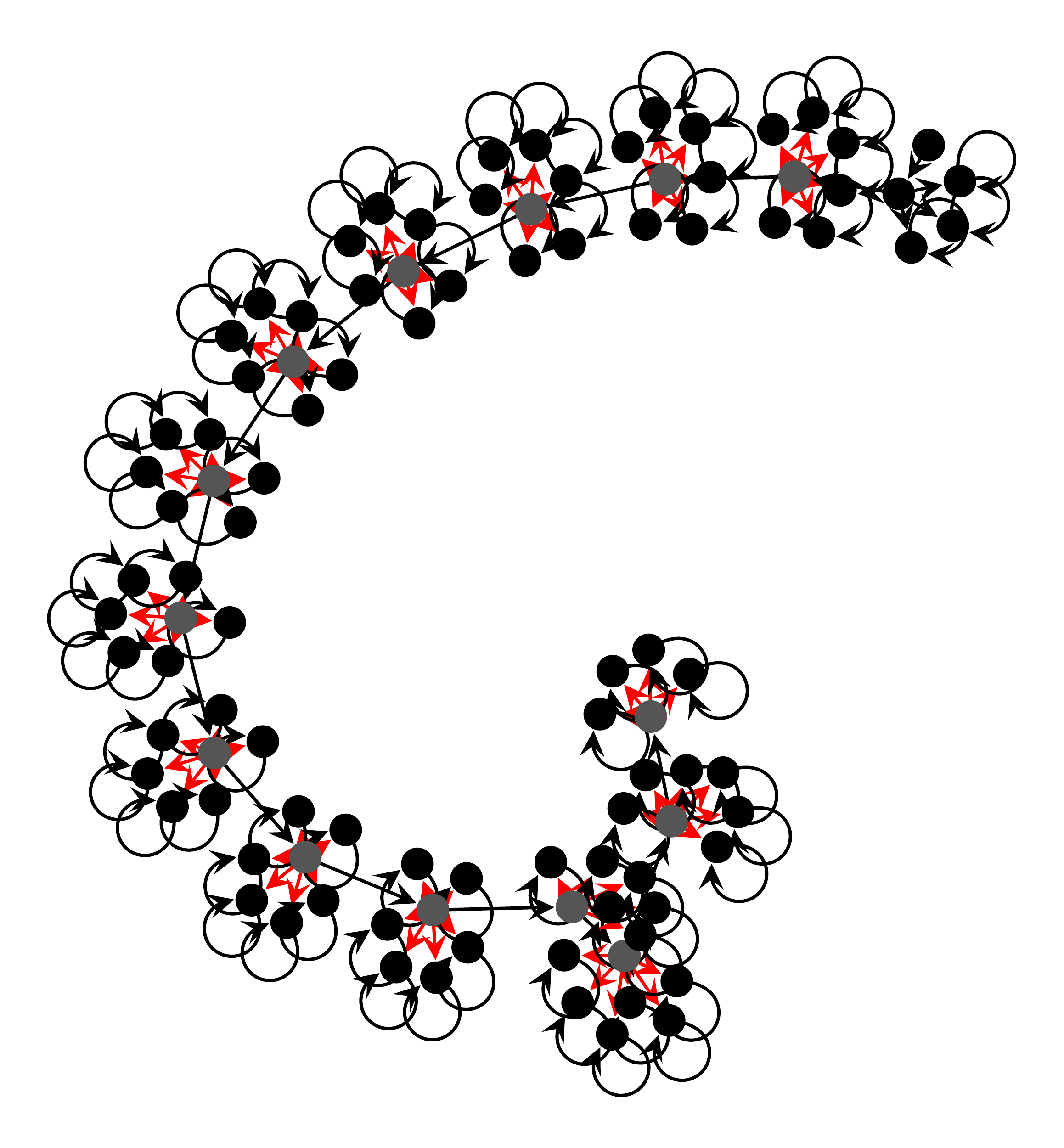}
      \includegraphics[width=0.45\textwidth]{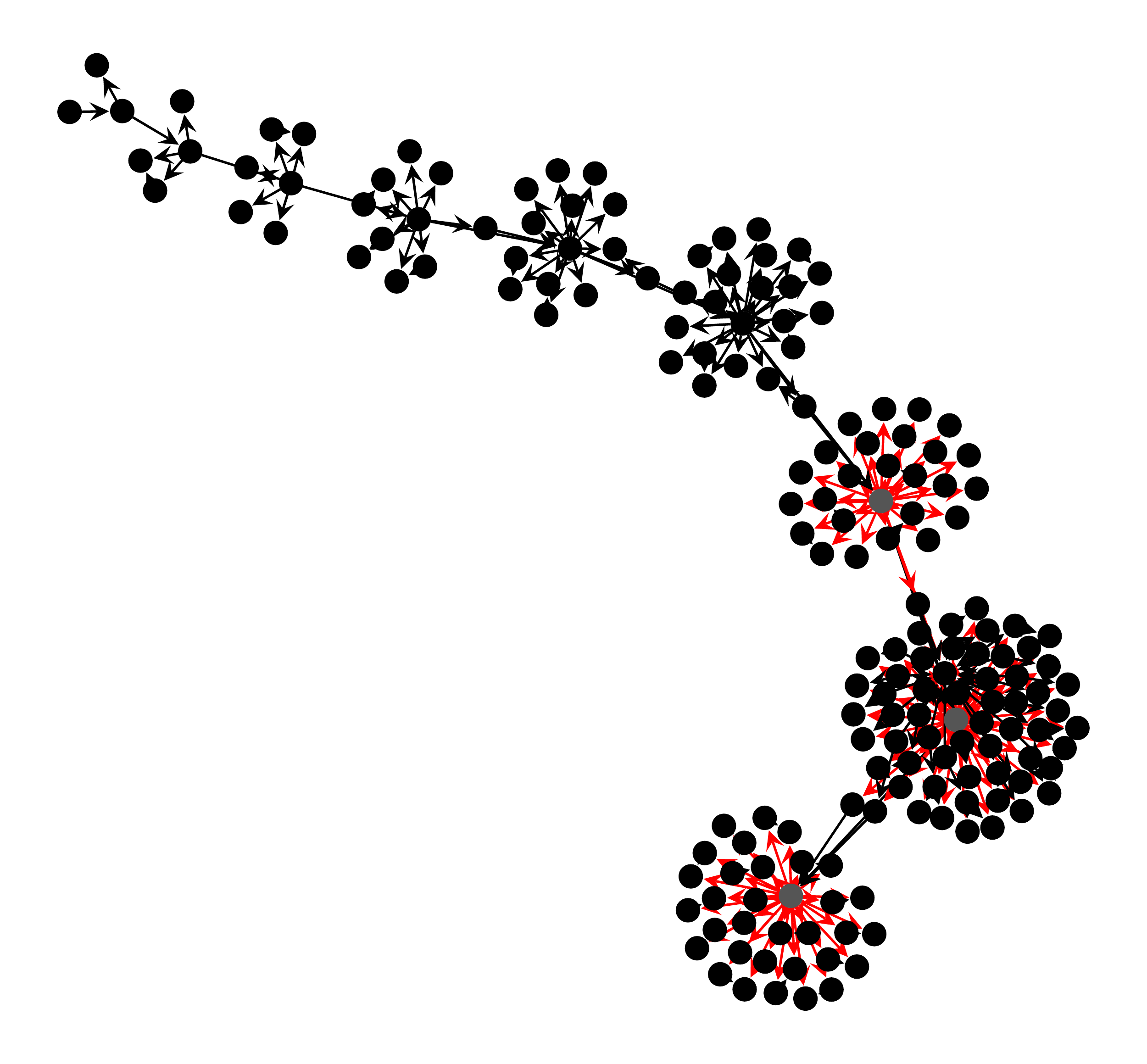}\\
      \includegraphics[height=0.45\textwidth]{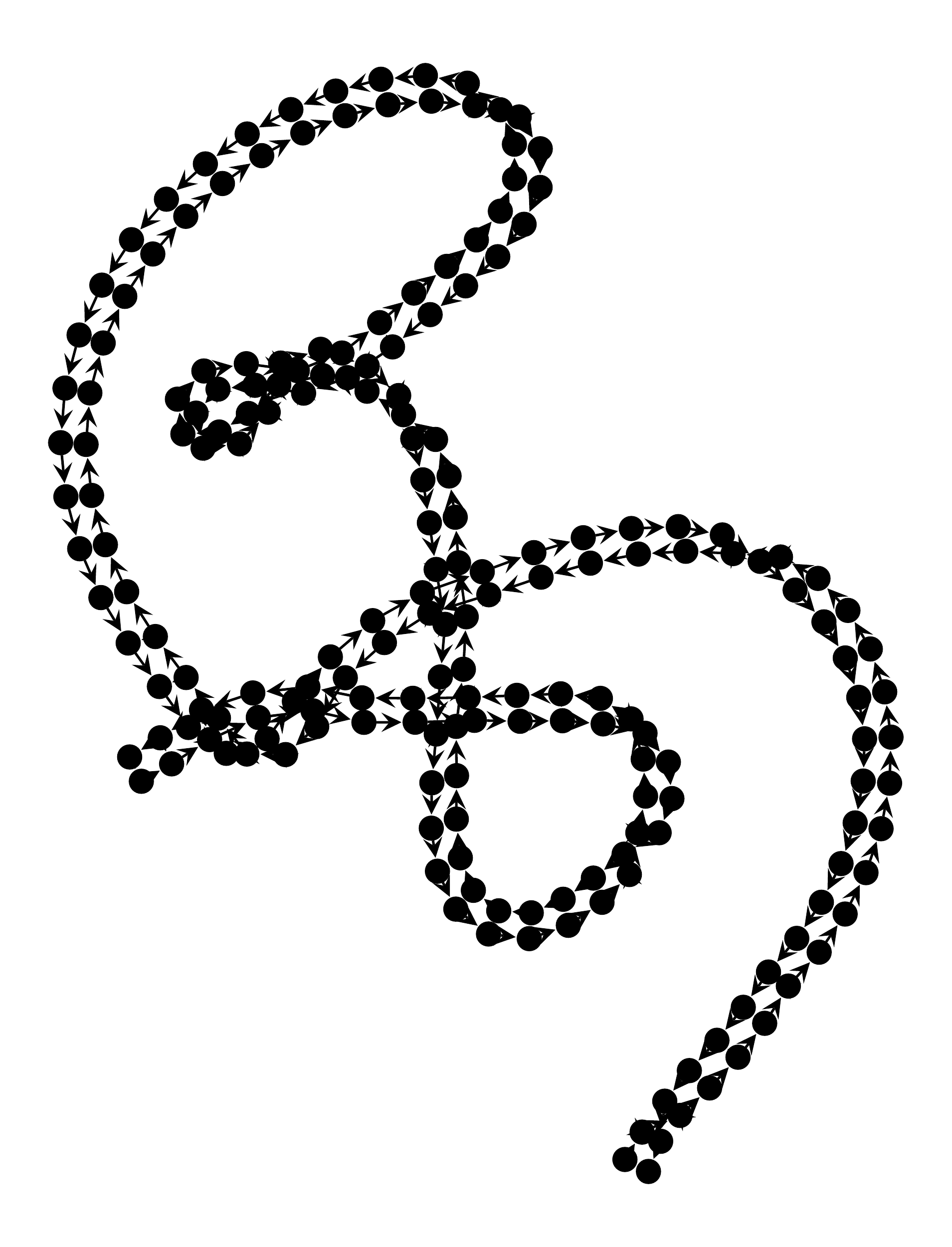}
    \end{minipage}
  }
  \hspace{0.4cm}
  \subfloat[Other]{
    \begin{minipage}{0.3\textwidth}
      \centering
      \includegraphics[width=0.45\textwidth]{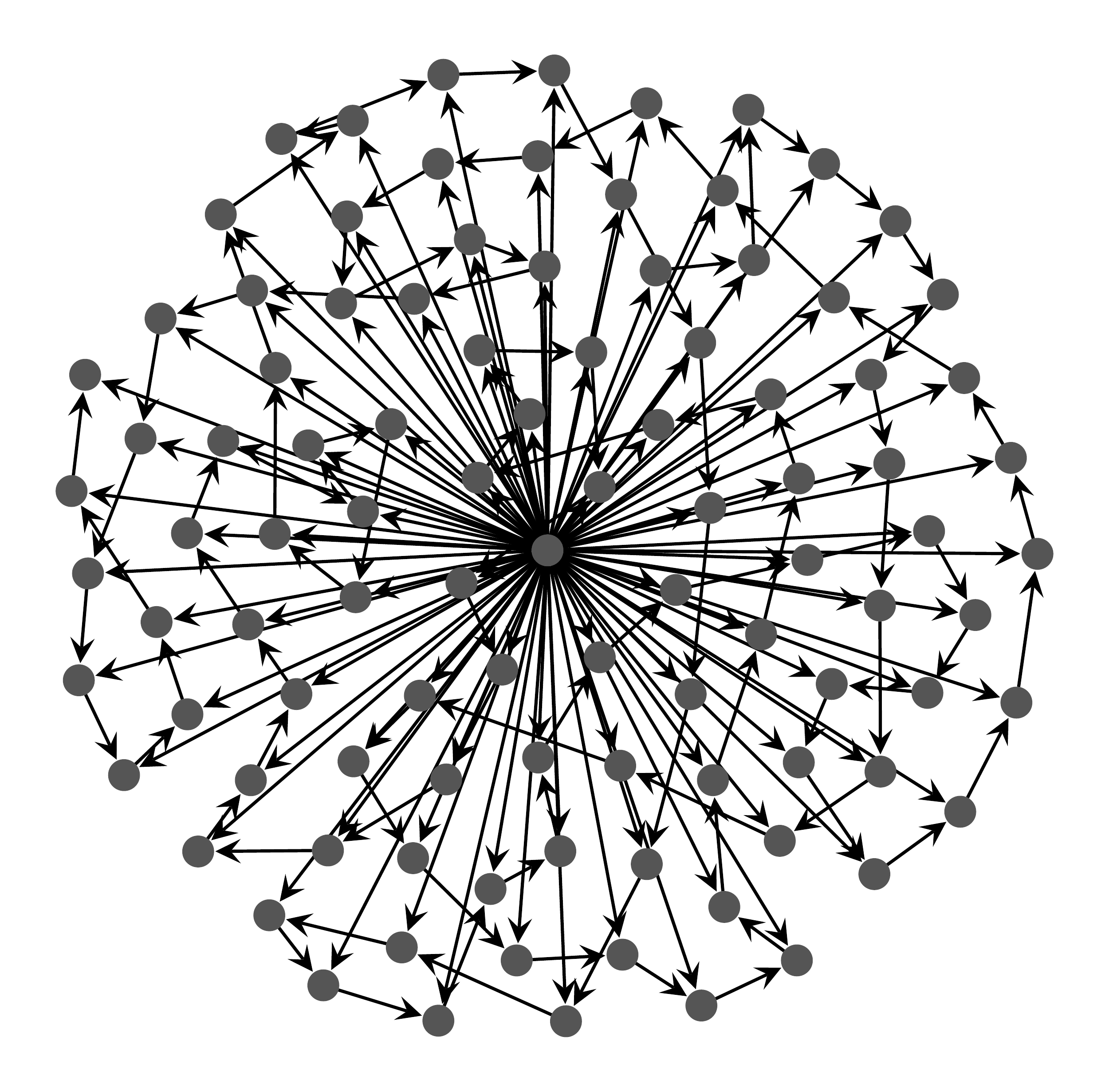}
      \includegraphics[width=0.45\textwidth]{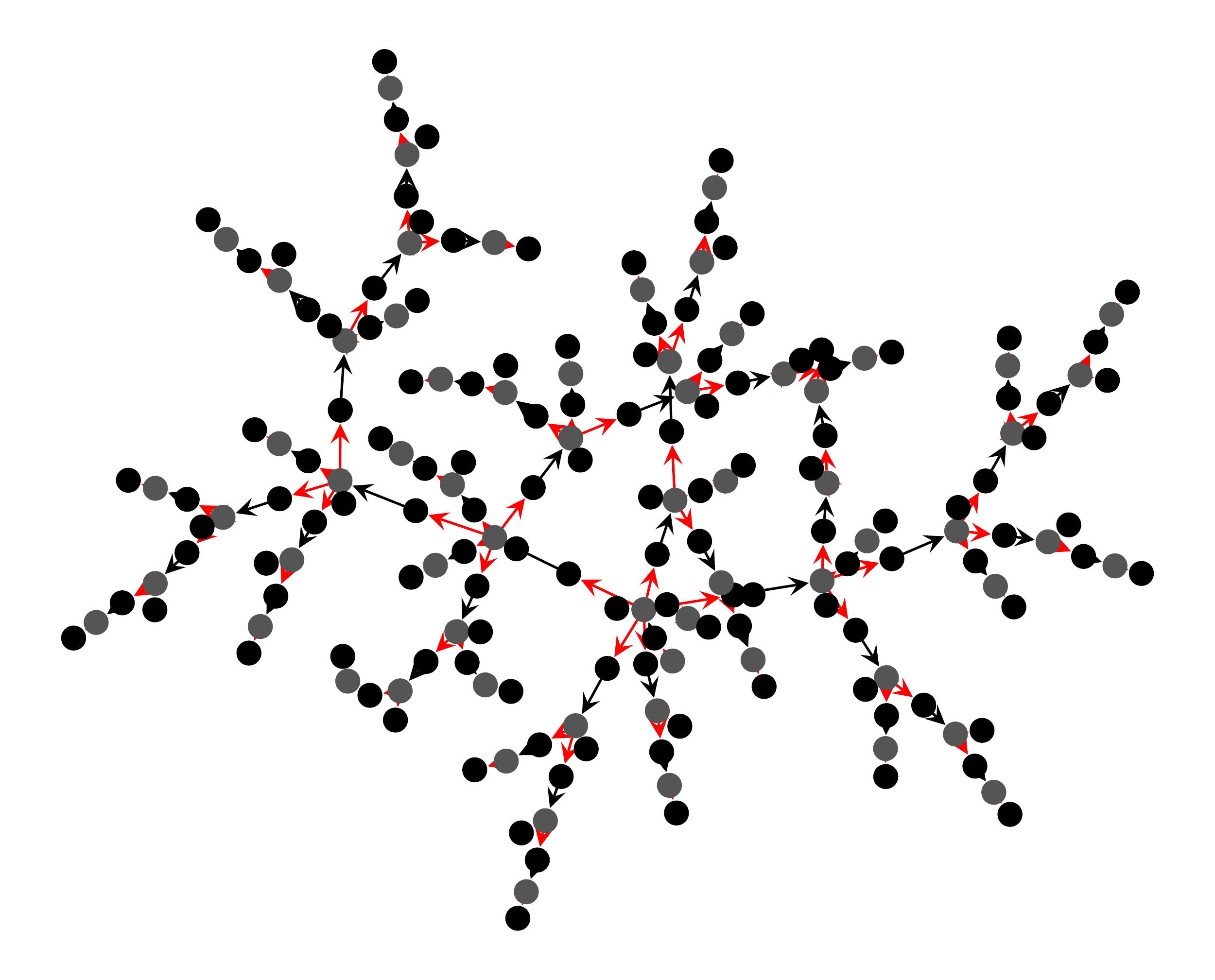}\\
      \includegraphics[height=0.45\textwidth]{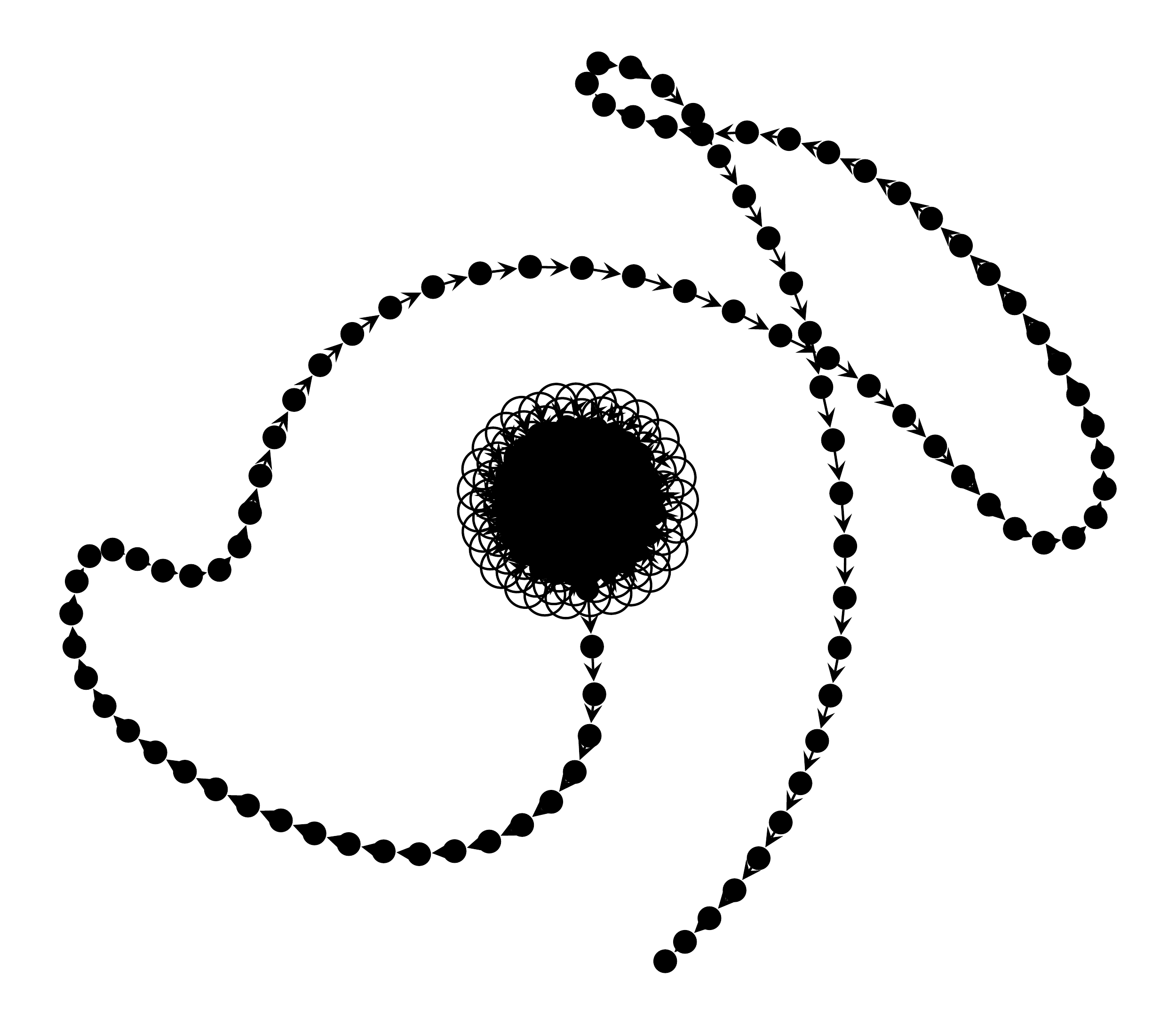}
    \end{minipage}
  }
  \caption{Samples of recurring structures. Reservoirs grown for the NARMA task exhibit ``life-like'' patterns. The diversity of these structures suggests that the DGCA model is capable of generating specialized networks with functional properties, reminiscent of natural systems.}
  \label{fig:structures}
\end{figure*}

We conducted two types of experiments: task-driven and task-independent growth. To calculate the fitness on the task-driven growth experiments, approximately two-thirds of the reservoir is perturbed with the input signal. Input weights are randomly generated from the discrete set \{$-1$, 0, $+1$\}, assigning each value with equal probability. For this study, the input gain and feedback gain are set to 0.1 and 0.95 respectively. 

Once the reservoir state is computed, the readout layer is trained using Bayesian Ridge regression between the state and target output. The regression is run for 3000 epochs, with a learning rate of $1\text{e}{-}6$. The reservoir’s performance is evaluated using the normalized root mean square error (NRMSE) between the true and predicted sequences. 

To mitigate variance caused by randomly-initialized input weights, each measurement is repeated five times. The model is also benchmarked against randomly-initialized reservoirs under similar constraints (binary weights and comparable density)---i.e.\ `normal' Echo State Networks (ESNs), to be used as a control case.

We chose the NARMA (Normalized Auto-Regressive Moving Average) family of tasks to drive the optimization. NARMA tasks are widely used as benchmarks for imitation of \textit{open} systems. These are based on two fundamental modeling approaches: Auto-Regression (AR) and Moving Average (MA). We will be using the generalized \(N\)-th order system, referred to as NARMA-\(N\) \citep{rodan_simple_2010}. As \(N\) increases, the reservoir must effectively retain and process a longer history of states to compute the output, making the task more challenging. The input remains in the range [0, 0.5].

In contrast, the task-independent growth experiments evaluated reservoirs using four widely-accepted RC metrics, as summarised in \citet{wringe2024reservoir}: Kernel Rank (KR), Generalization Rank (GR), Linear Memory Capacity (LMC), and Spectral Radius (SR). Rather than measuring performance on specific tasks, these metrics capture computational properties intrinsic to the reservoir. KR reflects the reservoir's ability to separate dissimilar inputs \citep{legenstein2007edge}; GR quantifies robustness to noise by measuring generalization over similar inputs \citep{legenstein2007edge,wringe2025reservoir}; LMC quantifies fading memory \citep{jaeger2001short}; and SR is often associated with a reservoir's operation near the edge of chaos \citep{jaeger2001echo}. For KR, GR, and LMC, the higher the metric, the more fit the reservoir is. For SR, fitness was evaluated from its distance to 1---the lower the distance, the better.

\subsection{Structures}

Many of the observed structures contained linear strands or strand-like characteristics. Figure~\ref{fig:structures} shows some sample reservoirs grown during the study, while illustrating the three categories into which the reservoirs were classified for subsequent analysis. Graphs are visualized using the graph-tool\footnote{See \url{https://graph-tool.skewed.de/}} library.

\begin{itemize}
    \item \textit{Linear} includes all graphs that form unbranched strands of nodes. Graphs with isolated chains (e.g.\ two disconnected strands) also fall under this category. 
    \item The \textit{Loosely Stranded} category includes all graphs that follow a strand-like structure but with slightly more complex patterns. These typically include double strands and strands with self-loops or cyclic subgraphs.
    \item \textit{Other} includes all remaining structures. These can be described as tree-like, clustered, hubs, among others. Unlike the previous two, no single dominant pattern emerged.
\end{itemize}

To systematically classify the graphs, we use a combination of graph-theoretic metrics. A graph is first treated as undirected, and divided into all its connected components. For each component, we compute the diameter. A component is identified as `linear' if it has a diameter of $n_c{-}1$ (where $n_c$ is the number of nodes in the component). 

If that condition does not apply, we consider whether the graph is `loosely stranded', measured by three conditions; if the normalized betweenness centrality for every component satisfies:

\begin{equation} \label{eqn:centrality}
\frac{g(v)}{(n_c - 1)(n_c - 2)} > 0.01
\end{equation}
where $g(v)$ is the betweenness centrality of node $v$, which quantifies the number of shortest paths between other node pairs that pass through $v$. This indicates that each node $v$ lies, on average, on more than 1\% of all such paths, suggesting that certain nodes consistently act as bridges within the network. Additionally, the reciprocal of closeness centrality, $C(v)$, must satisfy:

\begin{equation} \label{eqn:closeness}
\frac{1}{C(v)} > 0.07n_c
\end{equation}
where $C(v)$ is the inverse of the average shortest-path distance from node $v$ to all other nodes in the component. This indicates that, on average, nodes are at least $0.07n_c$ steps ($7\%$ of the component size) away from others, suggesting that the graph is relatively sparse or loosely connected. We also use the Gini coefficient of the component's node degrees to detect highly-connected nodes. The score must be under 0.1, indicating that there are no hub-like structures. We categorize graphs that do not meet any of the three conditions as `Other'.

\section{Results}

This section presents an analysis of the grown reservoirs and demonstrates how their performance varies based on budget and fitness functions. For each MGA run (i.e.\ 1000 iterations), the best-performing reservoir is selected based on the fitness function, resulting in 150 fittest reservoirs per experiment.

\subsection{Task-Driven Growth: NARMA-10}

These tests aimed to evaluate how the model compares to randomly-initialized (control) Echo State Networks (ESNs), under varying budget allocations for NARMA-10 growth. Here, \textit{budget} refers to the maximum number of permitted neurons in the reservoir. We integrated this into the MGA logic: if a grown reservoir exceeds the permitted size, it is assigned a fitness score of 0.

\begin{figure}[t]
    \centering
    \includegraphics[width=0.46\textwidth]{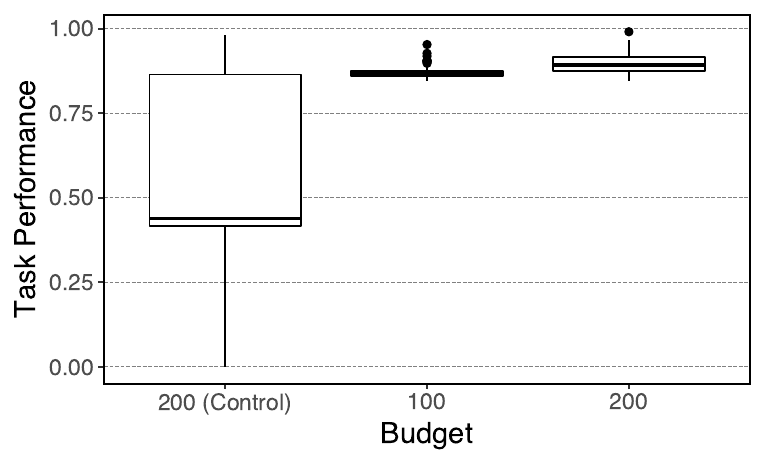}
    \caption{Task performance distribution of NARMA-10 grown reservoirs across node budgets. Pairwise comparisons used independent two-sample U-tests with Bonferroni correction ($\alpha = 0.05$, corrected threshold $p < 0.0167$). Significant differences: 100 vs 200 ($p=6.215\text{e}{-}20$), Control vs 200 ($p=1.852\text{e}{-}23$), Control vs 100 ($p=2.830\text{e}{-}15$).}
    \label{fig:narma10_perf}
\end{figure}

Figure \ref{fig:narma10_perf} shows the task performance (inverse NRMSE; higher is better) on NARMA-10 for the fittest reservoirs in three settings: Control (200), 100, and 200-node budgets. Growing the reservoir for the task consistently outperforms random initialization, even when the grown reservoirs have only half the neurons of the control. As expected, increasing the budget leads to improved performance, with a statistically-significant increase observed between 100 and 200 nodes.

\begin{figure}[t]
    \centering
    \includegraphics[width=0.46\textwidth]{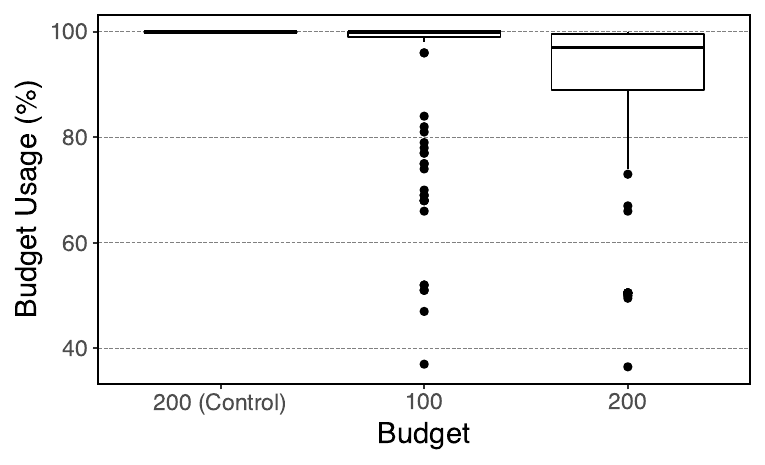}
    \caption{Budget usage (\%) distribution of NARMA-10 grown reservoirs across node budgets.}
    \label{fig:narma10_budget}
\end{figure}

Despite being given a budget, the DGCA model is not required to use all of it. Figure \ref{fig:narma10_budget} shows that many reservoirs did not use the full neuron budget, with some using less than 40\% (even in the 100-node experiment). Relative budget usage decreases when the budget increases. The control group remains fixed at 200, as its weights are randomly initialized for a fixed number of neurons.

\begin{figure}[t]
    \centering
    \includegraphics[width=0.46\textwidth]{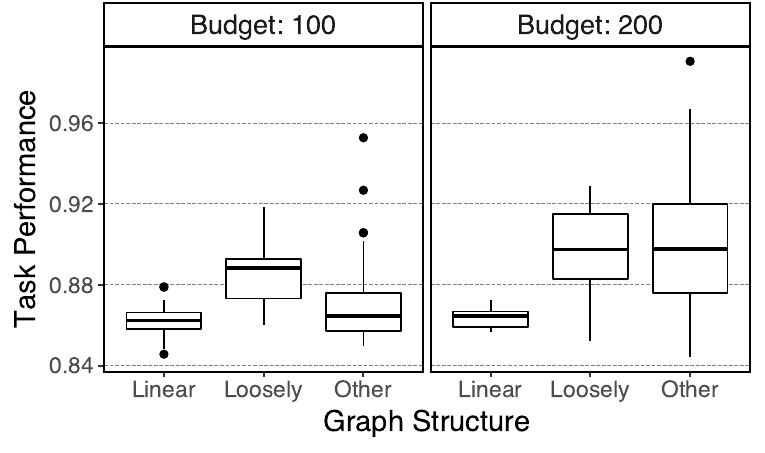}
    \caption{Structural distribution of task performance across node budgets. Pairwise comparisons used independent two-sample U-tests with Bonferroni correction ($\alpha = 0.05$, corrected threshold $p < 0.0167$). Significant differences, 100-node: \textit{Loosely Stranded} vs \textit{Linear} ($p = 1.240\text{e}{-}11$), \textit{Loosely Stranded} vs \textit{Other} ($p = 6.024\text{e}{-}5$). Significant differences, 200-node: \textit{Linear} vs \textit{Other} ($p = 1.041\text{e}{-}5$), \textit{Linear} vs \textit{Loosely Stranded} ($p = 7.206\text{e}{-}8$).}
    \label{fig:narma10_structures}
\end{figure}

Figure \ref{fig:narma10_structures} presents the distribution of task performance per structure of the fittest reservoirs. In the 100-node experiment, \textit{Loosely Stranded} graphs perform significantly better than the other two categories. In the 200-node experiments, \textit{Loosely Stranded} and \textit{Other} structures achieve the highest performance, while \textit{Linear} graphs consistently rank the lowest. These results suggest that under tighter resource constraints (e.g.\ reduced node budgets), \textit{Loosely Stranded} graphs provide an advantage.

\begin{table}[t]
    \centering
    \scriptsize
    \renewcommand{\arraystretch}{1.5}
    \begin{tabular}{cccccc}
        \hline
         \textbf{Exp.} & \textbf{KR} & \textbf{GR} & \textbf{LMC} & \textbf{SR} \\
        \hline
         \textbf{Control} & 0.025 (0.093) & 0.005 (0.010) & 0.025 (0.028) & 1.869 (1.631) \\
        \hline
         \textbf{100} & 0.620 (0.159) & 0.380 (0.259) & 0.490 (0.322) & 0.000 (1.000) \\
        \hline
         \textbf{200} & 0.609 (0.390) & 0.045 (0.160) & 0.077 (0.157) & 1.000 (1.000) \\ 
        \hline
    \end{tabular}
    \caption{Task-independent properties of NARMA-10 grown reservoirs. Values are reported as median (interquartile range). Each row corresponds to a NARMA-10 experiment with specific node budget, while each column corresponds to a metric computed from the fittest reservoirs. Kernel Rank (KR),  Generalization Rank (GR) and Linear Memory Capacity (LMC) are normalized by the number of nodes.}
    \label{tab:narma10}
\end{table}

 \begin{figure*}[t]
    \centering
    \includegraphics[width=\textwidth]{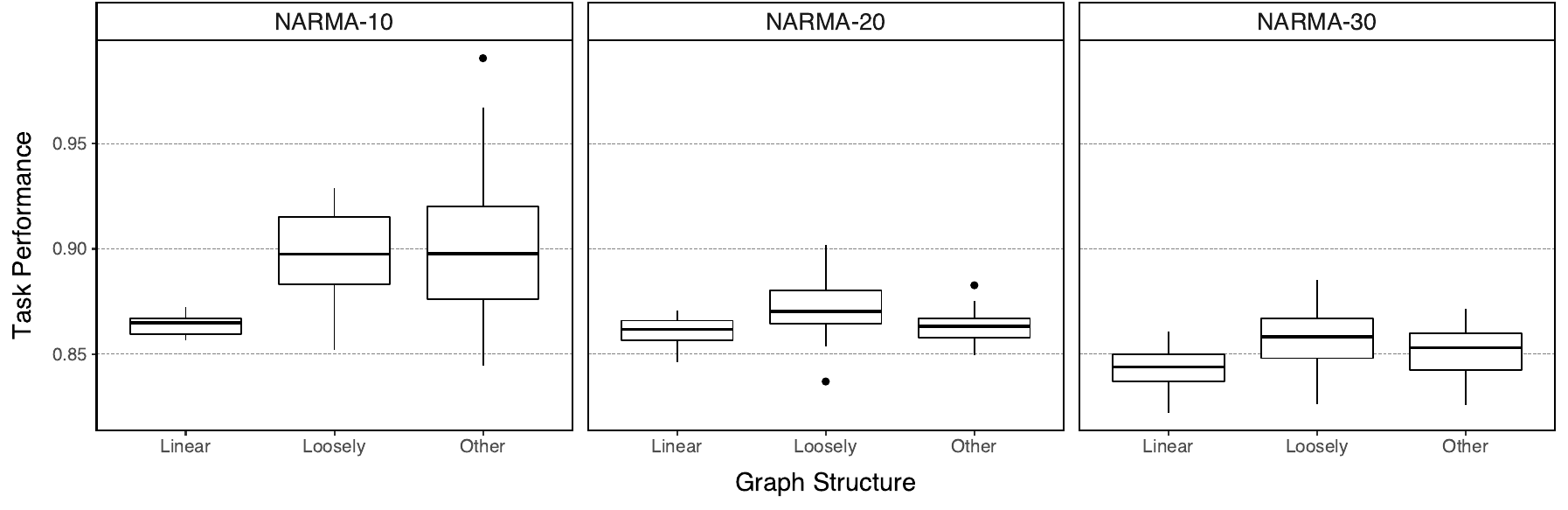}
    \caption{Structural distribution of task performance across NARMA orders. NARMA-10 results are reproduced from Figure~\ref{fig:narma10_structures} for reference. DGCAs are given a budget of 200 nodes. Pairwise comparisons used independent two-sample U-tests with Bonferroni correction ($\alpha = 0.05$, corrected threshold $p < 0.0167$). Significant differences, NARMA-20: \textit{Linear} vs \textit{Loosely Stranded} ($p = 7.206\text{e}{-}8$), \textit{Loosely Stranded} vs \textit{Other} ($p = 1.915\text{e}{-}4$). Significant differences, NARMA-30: \textit{Linear} vs \textit{Loosely Stranded} ($p=2.894\text{e}{-}6$), \textit{Linear} vs \textit{Other} ($p=2.929\text{e}{-}4$).}
    \label{fig:narman_structures}
\end{figure*}

Under constrained budgets, \textit{Linear} reservoirs are the most prevalent structure ($\approx$50\%). For larger budgets, \textit{Other} becomes the most prevalent category ($\approx$65\%). This suggests that a larger budget enables more structural diversity, while a smaller one favors simpler, reliable solutions.

Table~\ref{tab:narma10} presents four RC metrics of the fittest reservoirs. As expected, the control group's metrics are substantially worse than those of the task-grown reservoirs. The 100-node reservoirs show a considerably higher Generalization Rank (GR) and Linear Memory Capacity (LMC). It also has a median Spectral Radius (SR) of 0, which highlights the prevalence of linear strand structures, as these topologies have simple, non-chaotic dynamics. 

\subsection{Task-Driven Growth: NARMA-$N$}

\begin{table}[t]
    \centering
    \scriptsize
    \renewcommand{\arraystretch}{1.5}
    \begin{tabular}{cccccc}
        \hline
         \textbf{Exp.} & \textbf{KR} & \textbf{GR} & \textbf{LMC} & \textbf{SR} \\
        \hline
         \textbf{10} & 0.609 (0.390) & 0.045 (0.160) & 0.077 (0.157) & 1.000 (1.000) \\
        \hline
         \textbf{20} & 0.751 (0.260) & 0.242 (0.091) & 0.256 (0.136) & 1.000 (1.000) \\
        \hline
         \textbf{30} & 0.809 (0.174) & 0.247 (0.112) & 0.255 (0.133) & 1.000 (1.000)	 \\ 
        \hline
    \end{tabular}
    \caption{Task-independent properties of NARMA-$N$ grown reservoirs. Values are reported as median (interquartile range). Each row represents a different NARMA-$N$ experiment, while each column corresponds to a metric computed from the fittest reservoirs. Kernel Rank (KR),  Generalization Rank (GR) and Linear Memory Capacity (LMC) are normalized by the number of nodes.}
    \label{tab:narman_metrics}
\end{table}

\begin{table}[t]
    \centering
    \scriptsize
    \renewcommand{\arraystretch}{1.5}
    \begin{tabular}{cccccc}
        \hline
         \textbf{Exp.} & \textbf{KR} & \textbf{GR} & \textbf{LMC} & \textbf{SR} \\
        \hline
         \textbf{KR} & 0.784 (0.125) & 0.558 (0.477) & 0.292 (0.681) & 1.000 (1.000) \\
        \hline
         \textbf{GR} & 0.775 (0.059) & 0.588 (0.109) & 0.714 (0.162) & 0.000 (0.000)	 \\ 
         \hline
         \textbf{LMC} & 0.725 (0.085) & 0.588 (0.118) & 0.833 (0.173) & 0.000 (0.000) \\ 
         \hline
         \textbf{SR} & 0.136 (0.173) & 0.048 (0.038) & 0.026 (0.052) & 1.000 (0.000) \\
         \hline
         \textbf{All} & 0.811 (0.103) & 0.456 (0.142) & 0.595 (0.244) & 1.000 (0.000) \\ 
        \hline
    \end{tabular}
    \caption{Task-independent properties of metric-grown reservoirs. Values are reported as median (interquartile range). Each row represents a different task-independent experiment, while each column corresponds to a metric computed from the fittest reservoirs. Kernel Rank (KR),  Generalization Rank (GR) and Linear Memory Capacity (LMC) are normalized by the number of nodes.}
    \label{tab:metric_metrics}
\end{table}

To evaluate the adaptability of DGCA under increasing task difficulty, we grew a further 300 reservoirs with higher NARMA orders. We tested for $N{=}20$ and $N{=}30$ tasks with a fixed budget of 200 nodes. As expected, increasing task difficulty leads to a decrease in performance. Figure \ref{fig:narman_structures} shows performance distributions across NARMA orders and reservoir structures. For NARMA-20, \textit{Loosely Stranded} reservoirs outperform the other two categories, with no significant difference between \textit{Linear} and \textit{Other}. For NARMA-30, \textit{Linear} performs the worst, and \textit{Loosely Stranded} reservoirs exhibit the highest median performance.

Table~\ref{tab:narman_metrics} shows the RC metrics of the fittest reservoirs across NARMA orders. A median SR of 1 across all experiments indicates that \textit{Linear} structures are uncommon among successful solutions. GR and LMC are the only metrics that show substantial improvement when moving from NARMA-10 to the more difficult NARMA-20 and NARMA-30 tasks. This suggests that these two metrics may play a critical role in supporting reservoir performance under increasing NARMA complexity. Despite not explicitly optimizing for GR or LMC, the DGCA is able to specialize in ways that prioritize these metrics as task difficulty increases. 

\begin{figure*}[t]
    \centering
    \subfloat[ \label{fig:ind_vs_dep_a}]{
        \includegraphics[width=\textwidth]{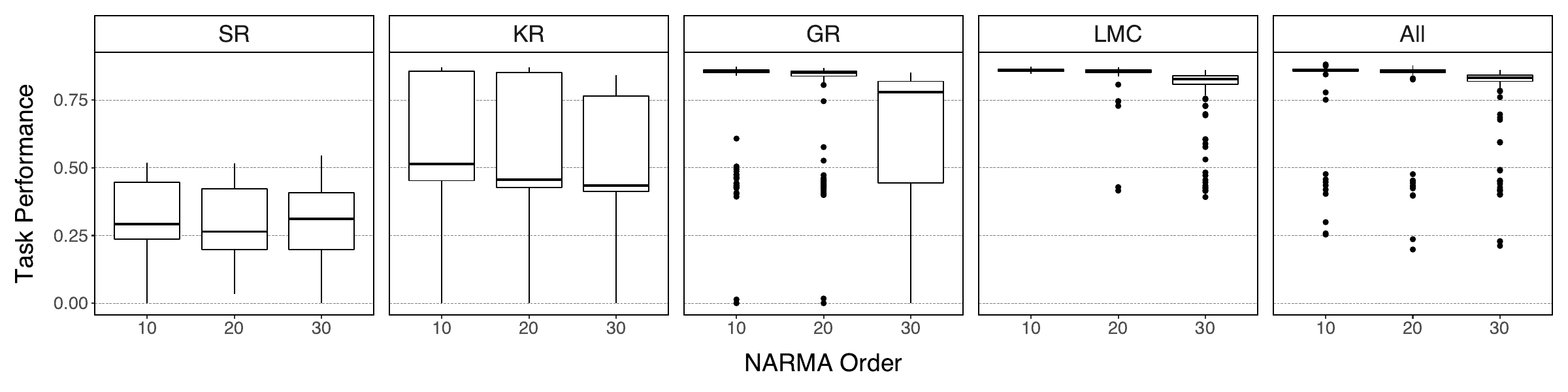}
    }
    \vspace{0.5em}
    \subfloat[\label{fig:ind_vs_dep_b}]{
        \includegraphics[width=\textwidth]{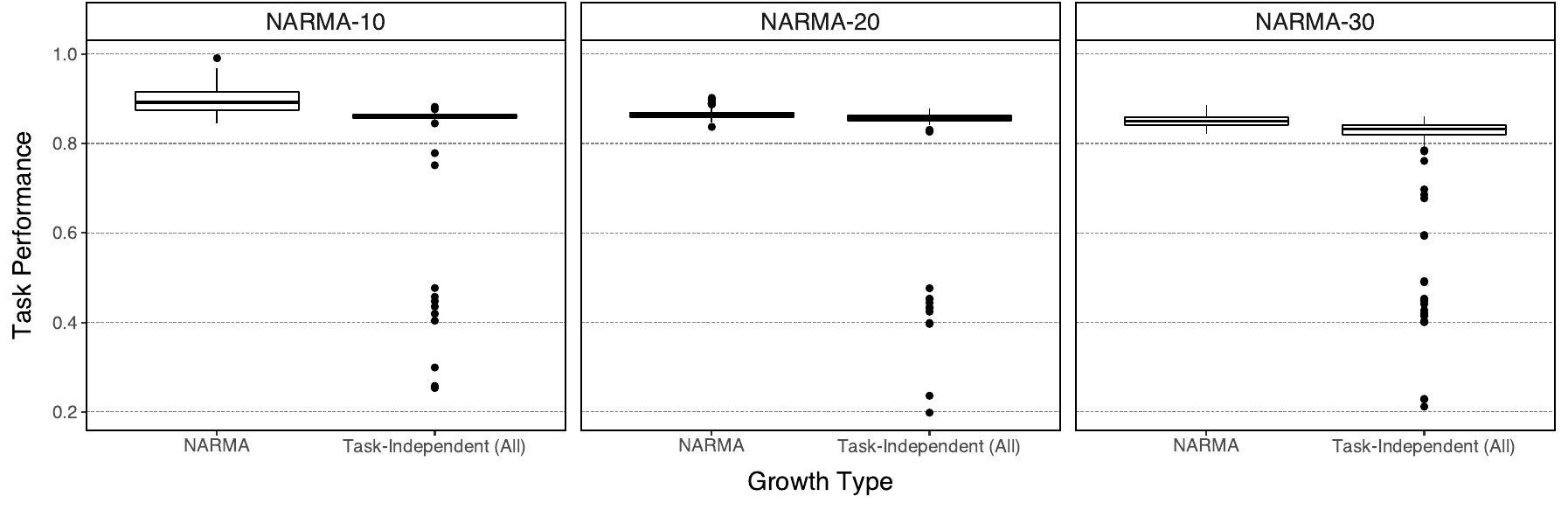}
    }
    \caption{Task performance distribution across growth type, NARMA orders and metrics. The chosen metrics are: Spectral Radius (SR), Kernel Rank (KR), Generalization Rank (GR), and Linear Memory Capacity (LMC). DGCAs are given a budget of 200 nodes. \textbf{(a)} Task performance comparison across metric and NARMA order. \textbf{(b)} Task performance comparison across growth types. Pairwise comparisons used independent two-sample U-tests with Bonferroni correction ($\alpha = 0.05$, corrected threshold $p < 0.0167$). Significant differences: NARMA-10 ($p=9.613\text{e}{-}36$), NARMA-20 ($p=1.994\text{e}{-}13$), NARMA-30 ($p=3.378\text{e}{-}23$).}
    \label{fig:ind_vs_dep}
\end{figure*}

\subsection{Task-Independent Growth}

\begin{figure}[t]
    \centering
    \includegraphics[width=0.46\textwidth]{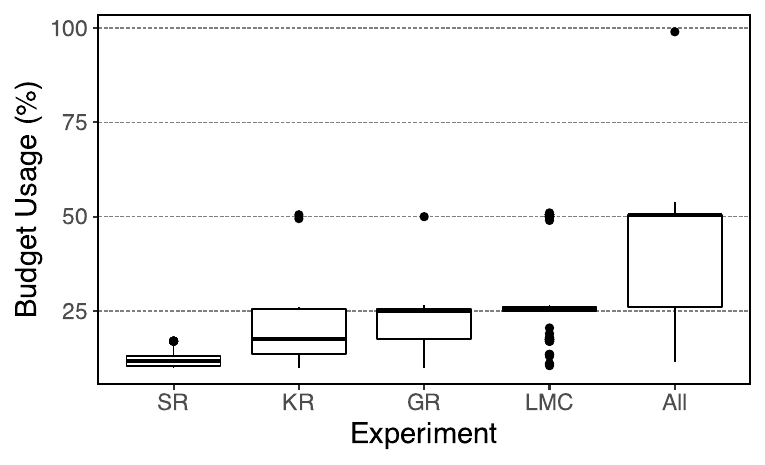}
    \caption{Distribution of budget usage (\%) across task-independent properties.}
    \label{fig:metric_budget}
\end{figure}

While we have shown that a reservoir can be grown to solve specific tasks by training directly on them, we now ask whether optimizing for task-independent properties (KR, GR, etc.) can yield reservoirs that generalize to unseen tasks. We set the DGCA’s fitness function to each metric individually, as well as to an additive combination of all four. All experiments were conducted with a 200-node budget.

Figure \ref{fig:ind_vs_dep_a} shows task performance across different target metrics and NARMA orders. As indicated in Table~\ref{tab:narman_metrics}, GR and LMC are the most effective approximators of task performance (after \textit{All}). SR is the worst-performing metric across all NARMA orders, followed by KR. The spread in task performance for KR-grown reservoirs is relatively high, suggesting that this metric is less robust than GR or LMC and may lead to unpredictable behavior in NARMA tasks.

Figure \ref{fig:ind_vs_dep_b} compares task performance distributions of task-grown (\textit{NARMA}) and metric-grown (\textit{All}) reservoirs. Note that the same task-independent reservoirs are used across all NARMA orders, whereas task-grown ones are tailored to each. Perhaps unsurprisingly, task-specific growth yields a significant performance improvement over the more general, metric-based approach. Notably, however, reservoirs grown using these metrics alone still perform well, despite the fact that such metrics are only indirect proxies of reservoir dynamics.

Figure \ref{fig:metric_budget} shows budget usage (\%) across the different metric-based experiments. In contrast to task-grown reservoirs, metric-grown reservoirs tend to use fewer neurons. The \textit{All} experiment uses the most ($\approx$50\%), while LMC uses only around 25\%.

Table \ref{tab:metric_metrics} shows the RC metrics of the fittest metric-grown reservoirs. GR and LMC experiments produce reservoirs `competitive' with task-grown ones. SR is 0 for both, suggesting that this metric is less relevant in the context of NARMA tasks. These scores, however, do not suggest that LMC or GR alone are exclusive predictors of NARMA performance, as earlier experiments showed that reservoirs with significantly lower memory capacity ($\approx$30\% of LMC) or generalization potential ($\approx$40\% of GR) were consistently able to outperform those with higher scores (see Table~\ref{tab:narman_metrics}).

\section{Discussion}

In this paper, we have demonstrated that DGCAs can be used to grow reservoirs. We have used an adaptation of the Microbial Genetic Algorithm (MGA) to search for optimal neural network weights based on fitness functions centered around task performance and RC metrics. The grown reservoirs were then benchmarked against imitation tasks of increasing difficulty. 

The structural diversity of the produced reservoirs indicates that distinct graph configurations can emerge as viable solutions to the same problem. Among these, some low-density, linear strand topologies emerged as the best performing. This finding suggests that even limited-resource, sparse networks can function as effective reservoirs. Moreover, recurring ``life-like'' structures proved to be consistently better adapted than others across experiments. These are, for all practical purposes, unachievable by the search of randomly-generated networks, as these often result in small-world reservoirs.

Growing reservoirs based on RC metrics alone demonstrated that some of these can serve as reasonable approximators of task performance. It is interesting, however, that task-grown reservoirs occasionally exhibit relatively high task performance despite having modest scores across all four inspected metrics. These counterexamples suggest that some of these standard RC metrics may not fully explain or predict task performance.

Furthermore, this work lays the foundation for the development of a DGCA capable of producing \textit{plastic} reservoirs. These reservoirs are particularly valuable in remote-location applications or embedded systems, where structural damage or changes in task demands may occur. In such scenarios, traditional fixed-structure reservoirs lack a recovery mechanism, requiring total replacement instead of self-repair or adaptation. Future work could explore artificial perturbations or task redefinitions to evaluate the system’s potential for continual learning.

Given that DGCAs were designed as a model for directed graph growth, the integration of a RC context also opens new avenues for research along these lines. We are looking towards integrating Liquid State Machines (LSMs) and a broader task domain.

Simulating physical reservoirs is a common practice in theoretical research, particularly in the early stages of developing ideas. At the experimental level, however, challenges remain in mapping RC models to physical systems. First, it may be unrealistic to perturb the entire system (i.e.\ input weights connected to all nodes). Second, it is often infeasible to observe the full state of the substrate. Inspired by physical reservoir computing, we have begun developing an input-output scheme that defines a fixed set of input and output nodes. In this setup, the reservoir grows around the fixed input and output neurons, similar to the use of probes in physical substrates \citep{cai2023brain}.

One of the disadvantages of using DGCAs to produce reservoirs is the slow training or optimization of neural network weights. Due to the lack of a differentiable update function, adjusting weights must rely on search-based methods rather than the more efficient gradient-based optimization. A promising avenue for future work is exploring \textit{Few-Shot} learning \citep{fei2006one}, which could enable the DGCA model to leverage prior training on benchmark tasks, allowing it to adapt more quickly to new, unseen tasks. We are also looking at the role of different activation functions \citep{griffin_evaluatingESNsagainst}, and how these might influence the behaviour of the grown reservoirs.

To conclude, we have demonstrated the ability of DGCAs to grow reservoirs. Our results, evaluated using task- and metric-based fitness functions, highlight the model's potential in reservoir computing. This work contributes to the advancement of DGCAs as a viable framework for reservoir computing and opens new directions for exploring growth mechanisms inspired by nature's purposeful growth.

% \newpage

\footnotesize
\bibliographystyle{apalike}
\bibliography{main} % replace by the name of your .bib file

\end{document}